%% file: arxiv-version-sparseapproxembeddings-5.tex
\documentclass[final,hidelinks,onefignum,onetabnum]{siamart220329}
\usepackage{enumitem,kantlipsum}

\input{sparseapprox_shared}

\ifpdf
\hypersetup{
  pdftitle={Network Embedding Using Sparse Approximations of Random Walks},
  pdfauthor={P. Mercurio, D. Liu}
}
\fi

\usepackage{booktabs}
\newcommand{\RR}{{\mathbb{R}}}

\newcommand{\ZZ}{{\mathbb{Z}}}

\theoremstyle{definition}
\newtheorem*{defn}{Definition}

\begin{document}

\maketitle

\begin{abstract}
In this paper, we propose an efficient numerical implementation of Network Embedding based on
commute times, using sparse approximation of a diffusion process on the network obtained by a 
modified version of the diffusion wavelet algorithm. The node embeddings are computed by optimizing 
the cross entropy loss via the stochastic gradient descent method with sampling of low-dimensional 
representations of green functions. We demonstrate the efficacy of this method for data clustering and 
multi-label classification through several examples, and compare its performance over existing 
methods in terms of efficiency and accuracy. Theoretical issues justifying the scheme are also 
discussed.
\end{abstract}



\section{Introduction}
Networks and graphs provide versatile and intuitive representations for many real world situations. A
wide variety of relational data, e.g., social networks, protein-protein interactions, bio-chemical reacting 
systems, citation networks, and many others, can be represented as graphs and networks. As a result, 
methods for analyzing these networks have consistently garnered much interest.  {\it Network 
Embedding},  or {\it Feature Learning for Graphs}, which provides a major framework for network 
analysis, seeks to map each node in a given graph to a point in a low-dimensional vector space such 
that proximity information from the original graph is preserved. In particular, the embeddings of nodes 
that are closely related in the original network through edge or path connections, mutual neighbors, 
overlapping proximities, will be close to each other in the embedding space. 

In many machine learning applications, the lower dimensional vectors representations are more 
convenient to exploit for downstream tasks than the original network. In this paper, we focus on node 
embedding methods, instead of embedding edges or whole networks, for their benefit of being more 
adaptive and applicable for a variety of practical situations such
as node classification, clustering, and link prediction \cite{survey2}. Node classification assigns class 
labels to unlabeled nodes based on samples of labeled nodes, while clustering algorithms group 
representations of similar nodes together in the target vector space, and link prediction is used to 
predict edges based on data containing edge information.

Most node embedding methods can be categorized according to the general techniques used to 
compute the embedding function including 1.) matrix factorization, 2.) random walk sampling and 3.) 
deep learning \cite{survey1}. Matrix factorization methods, such as Laplacian Eigenmaps \cite{lapeigs}, 
Locally Linear Embedding \cite{lle} and Graph Factorization \cite{ahmedetal}, involve factoring a matrix 
representation of the network (e.g. adjacency matrix, Laplacian, transition probabilities) by eigenvalue 
or singular value decompositions to determine embedding maps. Due to its ability to model nonlinear 
structural information, Deep Neural Networks have also been used recently in Network Embedding, in 
which the embedding and decoding functions consist of multiple stacked neural network layers. 

Random walk methods collect information about conditional probabilities of node observations by 
running short random walk trials on the network, then obtain embeddings via Stochastic Gradient
Descent (SGD) minimization of cross entropies. Like many of the monte carlo techniques, random walk 
methods apply to large high-dimensional networks with few limitations. In DeepWalk \cite{deepwalk}, 
based on the SkipGram language model \cite{Mikolov13E0,Mikolov13E1}, closely correlated nodes are 
embedded near each other according to the frequency of co-occurrence within unbiased, fixed length 
random walks on the graph. Node2Vec \cite{node2vec} replaces the simple unbiased random walks 
with walks controlled by adaptive parameters to either backtrack and stay near the starting node, or
explore higher degree neighbors. A recent generalization for directed networks was proposed in 
\cite{directedreps} with applications to the study of transition pathways of metastable chemical reacting
networks. 

In this paper, we propose an improved implementation of random walk methods for node embeddings
preserving the \emph{commute time} \cite{qiu}, based upon a sparse approximation of random walks on 
the graph. The approach achieves efficiency by avoiding Monte Carlo simulations of finite time random
walks, and takes into account statistics of paths of arbitrary length between different nodes. The 
{commute time} is defined as the expected time for a random walk to travel from one node $u$ to 
another node $v$, and then back to the starting node $u$. It can also be viewed as a time integration of 
the \emph{diffusion distance} introduced in \cite{coifman}, which is defined through the probability of the
random walk traveling from $u$ to $v$ in fixed time $t$. One advantage of utilizing the commute time, 
rather than the diffusion distance, is that it removes the need to choose an appropriate value for the 
parameter of time $t$. 

Numerically, the commute time between two nodes can be computed from the Green function, for which
using the Schultz method \cite{Schultz}, one can obtain an efficient approximation via an iterative 
product involving dyadic powers of  the random walk's transition matrix. The {Diffusion Wavelets} 
Algorithm (DWA) \cite{diffwavelets} offers one way to compute increasingly 
low-dimensional, compressed representations for these dyadic powers. By construction, at each scale
represented by a dyadic power of the diffusion operator, the DWA produces a basis of scaling functions
that span the same space as the eigenvectors corresponding to the largest eigenvalues. As a result, 
these scaling functions can be used to produce embeddings that are equivalent up to a rotation to those 
generated by kernel methods \cite{mahadevan}. 

Different from DeepWalk and Node2Vec, rather than directly minimizing the cross entropy, we begin 
by finding a sparse, low-dimensional approximation of the random walk. Specifically, we take inspiration
from diffusion wavelets, by adopting an alternative approach that numerically approximates the Markov
matrix and its powers with truncated Singular Value Decomposition (SVD) algorithms, from which we can
produce multiscale embeddings similar to the scaling functions generated
diffusion wavelets. To achieve robustness and overcome the error induced by the sparse approximation,
we introduce scaling parameters on each coordinate of the sparse approximation according to their 
contribution to the loss function. SGD is employed for the minimization of the cross entropy to 
obtain optimal embeddings. We show that such obtained embeddings compare favorably to those 
produced by existing methods, with classification accuracy measured by the F1 Macro scores. 

The rest of the paper is organized as follows. Section \ref{sec:background} provides some 
necessary background on random walk and SkipGram model based node embeddings, commute time 
embeddings and an introduction to random walk sparse approximation techniques. In Section 3, we
introduce the proposed method for learning Network Embeddings and its relationship to other kernel 
based embedding methods. In Section 4, we show that the algorithm using truncated SVD for 
computing the sparse approximation of the random walk improves performance and accuracy 
on larger graphs over the DWA implementation. Section 5 provides numerical 
experiments on several networks, comparing the proposed method with some existing feature learning
methods from the literature. Finally, section 6 addresses some theoretical issues regarding the scheme. 

\section{Network Embedding Methods based on Random Walks}\label{sec:background}
We first want to introduce basic notations and definitions of Network Embedding techniques based on 
random walks. The basic setup is an undirected graph $G(S,E)$, where $S$ denotes the set of nodes, 
and $E$ the set of edges. Generalization to directed graphs is straightforward. Let $|S|=N$ and 
$A^{N\times N}$ be the weighted adjacency matrix of the graph with weight $a_{ij}\ge0$ between nodes
 $v_i$ and $v_j$. The output will be a map 
\begin{equation}\label{eq:encode}
z_i=f(v_i): \ S\rightarrow\mathbb{R}^d, \quad\text{with}\ d\ll N.
\end{equation}
Certain proximities are preferred to be conserved. For example, edge weights $\{a_{ij}\}$ are also 
called first order proximities, since they are the first and foremost measures of similarity between two 
nodes. The vector $r_i=\{a_{ik}\}_{k=1}^{N}$  denotes the first order proximity between node $v_i$ and 
other nodes. The second order proximity between $v_i$ and $v_j$ can be determined by the similarity 
between $r_i$ and $r_j$, which describe the pair's neighborhood structures. Higher 
order proximities can be defined likewise. 

A general framework of Network Embedding consists of 4 major components: an encoding function, 
a decoding function, a loss function and a proximity function. The encoding function \eqref{eq:encode},
which maps nodes to vector, contains a number of trainable parameters that are optimized during the 
training phase, while a decoding function, which usually contains no trainable parameters, reconstructs
pairwise similarities from the generated embeddings. Once a proximity function is chosen for the 
network, a loss function that determines the quality of the pairwise reconstructions is minimized in order
to train the model. 

\subsection{Network Embedding using Monte Carlo simulations}
The class of algorithms compute vector embeddings by minimizing the loss function 
\begin{equation}\label{eq:cross}
L = \sum_{v_i} \sum_{v_j \in w(v_i)} -\log\left( P(v_j | v_i) \right),
\end{equation}
in which $v_i,v_j$ are nodes in the network, $w(v_i)$ represents a random walk of finite length starting
at node $v_i$, and $P(v_j|v_i)$ denotes the conditional probability that $w(i)$ will include node $v_j$. 
The SkipGram language model, originally proposed for word embedding of natural languages, adopts 
softmax units as the following:
\begin{equation}\label{eq:softmax}
 P(v_j | v_i) = \frac{  \exp(z_i \cdot z_j)  }{\sum_\ell \exp(z_i\cdot z_\ell)},
\end{equation}
where $z_k$ denotes the vector embedding for node $v_k$. The encoding function, referred as direct
encoding, is simply $z_i=Ze_i$, where $Z\in\mathbb{R}^{d\times N}$ contains the embedding vectors 
for all nodes $v_i\in S$ with $e_i$ being an indicator vector. The set of trainable parameters is the 
lookup matrix $Z$. For large networks with many nodes, this denominator in \eqref{eq:softmax} 
becomes numerically impractical. To overcome this, \emph{negative sampling} \cite{Mikolov13E1} 
replaces the softmax units with 
\begin{equation}\label{eq:negative}
P(v_j | v_i) = \sigma(z_i \cdot z_j  ) \prod_{\ell=1}^L \sigma(- z_i \cdot z_{\ell} ),
\end{equation}
where $\sigma(x) = 1/(1+e^{-x})$ and $L$ is the chosen size of negative samples drawn from the $3/4$ 
power of the unigram (empirical) distribution that allows less frequent words to be sampled more often.

The first SkipGram-based method for Network Embedding is DeepWalk \cite{deepwalk}, in which an 
unbiased random walk is simulated on the graph to generate $w(i)$ in \eqref{eq:cross}. Essentially, 
the walk has an equal chance of stepping to any node connected by an edge to the current node, since
the transition matrix is the row normalized adjacency matrix. SGD is then used to minimize the cross 
entropy \eqref{eq:negative} to obtain the embedding function. The Node2Vec \cite{node2vec} method 
adopts a similar approach, but with a slightly more complex choice of random walk. In particular, two 
parameters $p$ and $q$ are employed to bias the random walk simulations. The return parameter $p$ 
is used to control the probability that a random walk will backtrack to immediately revisit the previous 
node. The in-out parameter $q$ controls the likelihood that the walk will advance to a node within the 
one-step neighborhood of the current node that is different from the previously visited node. That is, for
larger $q$, the random walk will be more likely to explore parts of the graph that are further away from
the starting node. Notice that this random walk depends on the two previous steps, therefore is not 
Markovian. These parameters can be used to find the ideal balance between breadth-first sampling, 
which tends to stay nearer to the source node and get a more accurate picture of its immediate 
neighborhood, and depth-first sampling, which tends to visit parts of the graph further away from the 
source.  By finding optimal values of $p$ and $q$, one can engineer a new random walk that will provide 
superior node embeddings. The method we want to introduce here resembles that of Node2Vec in 
finding a modified of random walk that will lead to better node embeddings.

\subsection{Network Embeddings with commute times}
For a network $G(S,E)$,  with the adjacency matrix $A=(a_{ij})$, let the degree matrix $D$ be the diagonal matrix with 
$D(i,i) = d_i = \sum_j a_{ij}$. The volume of the graph will be given by $vol = \sum_i d_i$, and the 
\emph{random walk Laplacian} is defined as $L = I-D^{-1}A$. The \emph{normalized Laplacian} has the
form $\mathcal{L}=D^{1/2}LD^{-1/2}$ with eigendecomposition $\mathcal{L}=\Phi\Lambda\Phi^T$, 
where $\Lambda$ denotes the diagonal matrix of eigenvalues ordered so that 
$0=\lambda_1 \leq\lambda_2\leq...\leq \lambda_{N}$, and the eigenvectors are given by columns of 
$\Phi=(\phi_1| \phi_2|\ldots|\phi_{N})$. 

The discrete Green function is the left inverse of the operator $L$ and can be given as
\begin{align}
G(i,j)=\sum_{k=2}^{N}\frac{1}{\lambda_k}\left( \frac{d_j}{d_i} \right)^{1/2}\phi_k(i)\phi_k(j).
\end{align}
Here, we assume that the graph is connected. Otherwise $\lambda_k=0$ for some $k\geq 2$, 
and we can instead consider the Moore-Penrose pseudoinverse, $L^+ =  (L^TL)^{-1}L^T$, to which 
the methods and results can be easily extended.

We can further define the commute time $CT(i,j)$ to be the mean time for the Markov 
process on the graph prescribed by $L$ to travel from $v_i$ to $v_j$ and back, which 
satisfies
\begin{equation}\label{eq:ct}
CT(i,j)={vol}\cdot\left(\frac{G(i,i)}{d_i}+\frac{G(j,j)}{d_j}-\frac{G(i,j)}{d_i}-\frac{G(j,i)}{d_j}\right).
\end{equation}
It is shown in \cite{qiu} that the coordinate matrix for embeddings that preserve commute time 
is then given by $ \Theta = \sqrt{vol} \Lambda^{-1/2}\Phi^TD^{-1/2}$, which leads to 
$\Theta^T \Theta = vol \cdot GD^{-1}$. Thus, commute time embeddings are equivalent to those 
produced by the kernel PCA on a constant multiple of the matrix $GD^{-1}$.

The commute time embeddings are also closely related to DeepWalk embeddings. In 
\cite{networkembedsmatrixfact}, it is found that as the length of the random walks approaches infinity, 
the Network Embeddings produced from DeepWalk with adjacency matrix $A$ are equivalent to 
factorizing the matrix 
\begin{equation}\label{eq:1}
\Omega = \log \left( vol\left(\sum_{r=1}^w (D^{-1}A)^r \right) D^{-1}   \right)-\log(wL), 
\end{equation}
where the window size $w$ is the maximum steps for a given random walk trial to explore neighbors 
of a given node, and $L$ is the sample size of the negative sampling.
To see the connection between $\Theta$ and $\Omega$, we use the fact that 
$G = \sum_{k=1}^\infty T^k = \sum_{k=1}^\infty (D^{-1}A)^k$. As $w$ approaches $\infty$, then, the 
first term of \eqref{eq:1} approaches the elementwise logarithm of $\Theta^T \Theta$. Commute time 
embeddings can therefore be interpreted as the elementwise exponential of a DeepWalk embedding
with infinite context size.
   
\section{Network Embedding Methods  based on Sparse Approximations}
We now propose a method to produce Network Embeddings that asymptotically preserve commute 
times while minimizing the cross entropy loss. The idea is to first obtain a sparse approximation of the
random walk on graphs with a multiscale approach, which will be used as a basis for approximating 
commute times. The embeddings are then generated by minimizing cross entropy loss with respect to
weights on these basis vectors using SGD.

Let $T=D^{-1}A$ denote the transition matrix for the random walk on a network with $N$ nodes. We
here assume that $T$ is local in the sense that its columns have small support, and that high powers of 
$T$ will be of low rank. We start by computing the compressed representations $T_k$ of the dyadic 
powers of $T$, i.e. $T^{2^k}\approx T_k$. To do this, we apply an orthogonalization procedure, e.g. 
DWA or truncated SVD, to the columns of $T$, producing an orthogonal basis 
$U_1 = \{u_1,u_2, \dots u_M\}_{\{M\le N\}}$, that approximates the range of $T$.  Then, we can write 
$T_1 = U_1^T T^2 U_1$, so that $T_1$ is an approximation of $T^2$ expressed in terms of the basis 
$U_1$. We repeat the process, orthogonalizing the columns of $T_1$ to produce an approximate basis
$U_2$ for its range, using this new basis to compute a representation of 
$T_2 =U_2^TT_1^2U_2 \approx T^4$, and so on. In \cite{diffwavelets, mahadevan},  scaling functions of
diffusion wavelets are adopted as basis of the span of the columns of $U_k$. We will
introduce a procedure of truncated SVD for for the basis. Details of implementation and numerical 
comparision will be provided in later sections. 

The unnormalized Green function $G$ can be computed from $T_1,\dots, T_k$ using the Schultz 
method, which iteratively computes the Green function (or Moore-Pembrose pseudoinverse) associated 
with  $L$ through $G_{K+1} = G_K ( 2I-LG_K)$. Using the fact that $L = I-T$, and taking $G_0 = I+T$, 
we obtain $G_{K+1}= \prod_{k=0}^K (I+T^{2^k})$. The Green's function ${G}$ then has the following
form on the complement of the eigenspace for the eigenvalue equaling 1, as also shown in 
\cite{diffwavelets}:
\begin{equation}
{G} = \sum_{k=1}^\infty T_k  =  \prod_{k=0}^\infty(I + T^{2^k})\approx G_K,
\end{equation}
which in turn allows for the quick computation of the commute times via formula \eqref{eq:ct}.
Network Embeddings that asymptotically preserve commute time relationships therefore can be 
generated with  the embedding matrix $\tilde\Theta$ satisfying 
\begin{equation}
\tilde\Theta^T\tilde\Theta=vol\cdot G_KD^{-1}.
\end{equation} 
In particular, since taking powers of a Markov matrix is equivalent to running the Markov process 
forward in time, we therefore obtained multiscale embeddings by  repeating the same approximation 
process for compressed representations of the Green function. 

While the matrix $\tilde\Theta$ can itself be used as an effective node embedding, we can further 
improve our results by using this sparse approximation as a starting point to obtain node embeddings 
that minimize the cross entropy.  The commute time based embedding using sparse approximations
results in a basis spanning the range of the sparse approximation to the powers of the random walk. 
We can re-weight the vectors in the same basis such that cross entropy loss is 
minimized.  Given the $M\times N$ sparse approximation $\tilde\Theta$, where the $i$th column 
$\tilde\Theta_i$ provides an embedding vector of node $v_i$, we aim to find constants 
$C_1, C_2, ... C_M$ such that the embedding given 
by $C\tilde\Theta$ where $C=diag(C_1,C_2,...C_M)$ minimizes cross entropy \eqref{eq:cross}.
Using SGD, we choose the optimal constants so that the role of the more ``important" columns
corresponding to particular singular values of the approximation to the random walk will be emphasized. 

Additionally, a residual correction technique is employed such that with some probability, singular 
vectors that were removed in the previous truncations can be reintroduced for the robustness of the 
algorithm to recover information lost in the approximation steps. Reintroducing columns from previous 
steps is also analogous to multiscale iterations between different time scales. At each step of SGD, 
with some small probability $\delta$, we also extend the basis $\tilde\Theta$ with one of the indices from
the basis in previous steps. That is  we randomly select a column $u_j$ of $U_{K-1}$, then project 
$u_j$ onto the basis $U_k$, and append the transpose of this projection to $\tilde\Theta$ as the bottom
row. This allows for the reintroduction of some information from the high-pass filtering 
subspaces, while still benefiting from the low-dimensional compression given by the low-pass bases.

\begin{algorithm}[t]
\label{alg:sgd}
\caption{Implementation of SGD minimizing cross entropy loss, over a sparse approximation of a 
diffusion process as input. At each step, with probability $\delta$, vectors from coarser scales are 
appended to the embeddings.}
\begin{algorithmic}
\STATE{\bf Input} $M$-scale embedding matrix $\tilde\Theta \in \RR^{M \times N}$, 
parameters $\delta\in(0,1)$.
\STATE{\bf Result}{ Updated $\tilde\Theta$ such that cross entropy loss is minimized.} 
\\ \ \vspace{-.1in} \\
{Initialize $C$ as the $M \times M$ identity matrix.\;}
\FOR{$1 \leq i \leq M$}
	\FOR{$j \in$ batch}
		\STATE{With probability $\delta$:, select a vector $u$ from the vectors $U_{K-1}$}
		\STATE{$\tilde\Theta = append( \tilde\Theta, U_kU_k^Tu),\quad C = diag(C, 1) $}
		\STATE{$\frac{dL}{dC} =\frac{d}{dC}L(C\tilde\Theta)$}\\			
		\STATE{$C = C - \diag(\eta*\frac{dL}{dC})$ }
	\ENDFOR
\ENDFOR
\RETURN $C\tilde\Theta$ 
\end{algorithmic}
\end{algorithm}

\section{Numerical Details of Sparse Approximation}

As described before, the sparse approximation process we apply to the random walk $T$ yields at each 
scale a basis of  scaling functions $U_k$ that spans an approximation of the range of 
$T^{2^k}\approx T_k$.  
Then, both representations of $G$ and node embeddings can be constructed for each compression 
level $k$ (with $k=1$ representing the finest scale, and larger $k$ representing increasingly coarse 
scales) from the scaling functions.  Essentially, once we have the sparse approximation of 
$G\approx G_k$, to obtain the $k$-level embedding for node $x$,  we extend the SVD at the $k$th level,
$vol\cdot G_kD^{-1} \approx \tilde{U}_k^T \tilde{\Sigma}_k \tilde{U}_k$, to the original basis and take the
matrix $\tilde\Theta = \tilde{\Sigma}_k^{1/2}\tilde{U}_k$ to be the embedding matrix; that is, the $n$th 
row of $\tilde\Theta$ will provide the embedding coordinates for the $n$th node. 

We now elaborate on two algorithms for these sparse approximations of the random walk. 
First, we give an overview of the diffusion wavelet algorithm introduced in \cite{diffwavelets}, and then 
propose an alternate implementation using a truncated SVD.

\subsection{Diffusion wavelets algorithms} 
Diffusion wavelets \cite{diffwavelets} extends ideas from wavelet theory to a more general setting. 
In particular, it extends the descriptions of wavelet bases in \cite{wavelets} for multiscale analysis of 
networks. In the diffusion wavelets setting \cite{diffwavelets}, the multiscale analysis includes a 
sequence of subspaces $V_0 \supset V_1 \supset V_2 \supset \cdots  $ 
such that $V_0$ is the range of the operator $T$, and for $m \in \ZZ$, $V_m$ represents an
approximation of the range of $T$ at scale $m$. In particular, when $\{ \zeta_i \}_{i \in \ZZ^+}$ and 
$\{ \lambda_i \}_{i \in \ZZ^+}$ denote respectively the eigenvectors and eigenvalues of $T$, $V_m$ is
defined as the span of $\{ \zeta_\lambda : \lambda^{2^{m+1}-1} \geq \varepsilon \}$. Each $W_m$ can
then be defined as the orthogonal complement of $V_m$ in $V_{m-1}$, as in the original wavelet setting.
Note that the parts of the spectrum with $\lambda^{2^{m+1}-1} \geq \epsilon$ can be viewed as 
"low-pass", so that the $V_m$ correspond to the ``approximation" or ``smoothing" subspaces in
classical wavelet theory, and the $W_m$ correspond to the "detail" components. 

The scaling functions $\Phi_m$ at each level $m$ then provide a basis for $V_m$, approximating a 
dyadic power of $T$, and is constructed by orthonormalizing the columns of the representation of $T$ 
at the previous level, $T_{m-1}$, up to precision $\varepsilon$. In fact, the subspace spanned by these 
scaling functions is an $\varepsilon$-approximation of $V_m$, which is denoted as $\tilde{V}_m$. As we
will see, $T_{m}$ is a representation of $T_{m-1}^2$ by construction, so each step of the diffusion 
wavelets algorithm dilates the scale by a factor of 2.

The Diffusion Wavelets Algorithm (DWA) can be outlined as follows: As inputs,  takes the operator 
$T=T_0$ and a parameter $K$ determining the number of iterations to be performed. The columns of 
$T$ are originally expressed in the basis $\Phi_0=\{ \delta_i \}_{1 \leq i \leq N}$. First, the columns of $T$ 
are orthonormalized up to $\epsilon$ using a rank-revealing $QR$ algorithm or a  modified 
Gram-Schmidt process with pivoting. This results in an orthonormal matrix that gives an 
$\varepsilon$-span $\Phi_1$ of the columns of $T$ written in terms of the original basis $\Phi_0$, 
which suggests a coordinate change matrix $[\Phi_1]_{\Phi_0}$.  $\Phi_1$ forms a basis for the 
subspace $\tilde{V}_1$,  i.e., the range of $T$, as $\epsilon$-approximation of $V_1$. We also obtain 
from the $QR$ decomposition an upper triangular matrix representation of $T$, denoted 
$[T]^{\Phi_1}_{\Phi_0}$, expressed as coordinate change from $\Phi_0$ in the domain to $\Phi_1$ in 
the range.

Then, the product 
$[\Phi_1]_{\Phi_0} [T^2]^{\Phi_0}_{\Phi_0} [\Phi_1]_{\Phi_0}^* =  [T^2]_{\Phi_1}^{\Phi_1}=:T_1 $ 
provides a representation of $T_0^2$ in terms of  $\Phi_1$. It is also possible, if desired, to compute a 
basis of wavelet functions spanning the orthogonal complement of $\tilde{V}_1$ in $\tilde{V}_0$, 
denoted $\tilde{W}_1$, by orthonormalizing the columns of $I-[\Phi_1]_{\Phi_0}[\Phi_1]_{\Phi_0}^*$. By 
iterating this entire process, we can compute representations of 
$T_k=[T^{2^k}]_{\Phi_{k-1}}^{\Phi_{k-1}}$, scaling functions $[\Phi_k]_{\Phi_{k-1}}$ spanning the space 
$\tilde{V}_k$ that $\varepsilon$-approximates $V_k$ (i.e. the range of $T_k$), and wavelet functions 
$\Psi_k$ for levels of various $k$ values.

\subsection{Sparse approximation via SVD}
We want to propose an SVD-based multiscale algorithm for approximating the dyadic powers of the 
transition matrix and the Green function, which proves to be more efficient and accurate, when applied
to the commute time based Network Embedding for node classifications, compared with DWA .
More specifically, at each step we compute a truncated SVD of $T_k$. The singular vectors associated 
with the largest singular values then form a basis spanning approximately the range of $T_k$. We can 
then use these singular vectors to compute a compressed representation of the next dyadic power, 
$T_{k+1} \approx T^{2^k}$, with respect to the basis $U_k$. 

First, let $T=U\Sigma V^T$ denote the full SVD of $T$, so that $U=U_0$ is the original basis.
We compute a truncated SVD of $T$, retaining the $j$ largest singular values and the corresponding 
singular vectors, so that $T \approx U_1 \Sigma_1 V_1^T$ where $U_1 \in \RR^{N \times j}$, 
$\Sigma_1 \in \RR^{j \times j}$, and $V_1^T \in \RR^{j \times N}$.
Then, $U_1$ provides an orthogonal basis for approximately the range of $T$, for some
margin of error determined by the choice of $j$. To obtain a representation of $T^2$ with
respect to the basis $U_1$, we use $T_1:=U_1^T T U_1$. 
To justify this, note that the left singular matrix $U$ has orthogonal columns, and furthermore 
these columns span the range of $T$. In particular, the left singular vectors corresponding to
the largest singular values span an approximation of the range of $T$, and therefore they play a role 
analogous to the diffusion wavelet scaling functions.

To obtain representations for
$T^{2^k}$ for $k=2,3,...,$ we can simply iterate this process by finding the truncated SVD of
$T_{k-1}$, taking $U_k$ to be the left singular matrix of that decomposition, and setting $T_k =
(U_{k}^T T_{k-1} U_k)^2$. Additionally, each $U_k$ can be represented in the original basis using
the change-of-basis formula: 
\begin{align}\label{eq:chain}
\left[U_{k}\right]_{U_{0}}=
\left[U_{1}\right]_{U_{0}} \cdots\left[U_{k-1}\right]_{U_{k-2}}\left[U_{k}\right]_{U_{k-1}},
\end{align}
where $\left[U_{k}\right]_{U_{k-1}}$ represents $U_k$ expressed in terms of the columns of
$U_{k-1}$, and $U_0\in \RR^{N \times N}$ represents the original basis of the columns of $T$. 
Note that in this notation, $\left[U_{k}\right]_{U_{k-1}}=U_k$.

The performance advantage of this implementation is demonstrated on the Butterfly dataset, for which
more details and numerical experiments can be found in section \ref{sec:numerical}. 
The 3D embeddings produced via the SVD algorithm perform better than both the 3D diffusion wavelet
embeddings and the commute time embeddings described in \cite{qiu} using 3 first columns. 
Additionally, the SVD approach allows more control over the rate of dimensional reduction, since one 
can directly choose the proportion of singular values to be retained. 
\begin{table}[h!]
  \begin{center}
    \caption{Comparisons between three-dimensional diffusion wavelet-based embeddings, truncated SVD-based 
    embeddings, and the commute time embeddings introduced in \cite{qiu}, on the butterfly dataset \cite{butterflydata1, 
    butterflydata2}. Runtime represents the time required to compute sparse approximations for powers of $T$ and $G$ 
    via the given algorithm. }
    \label{tab:svdcomparison}
    \begin{tabular}{lcr}\toprule 
      \textbf{Method} & \textbf{Runtime} & \textbf{F1 Macro}\\
      \midrule
      Diffusion Wavelets & 70 s & 0.5827 \\
      SVD  &  {3.5 s} & {0.7490}  \\ 
      Commute Time Embeddings &  0.5 s & 0.6051\\
     \bottomrule
    \end{tabular}
  \end{center}
\end{table}

In the case where the random walk matrix $T$ is symmetric, the connection to the diffusion wavelet 
based algorithm is obvious; the singular values are the absolute values of the eigenvalues of $T$, and
therefore the largest singular values are the eigenvalues with largest-magnitudes. The columns of the
left singular matrix $U$ then give the eigenvectors of $T$.  If we select $j$ such that $\forall s \leq j$, 
$\lambda_s^{2^{m+1}-1} \geq \varepsilon$, then the first $j$ columns of $U$, when the singular values 
are ordered from largest to smallest, must span  $V_m := \langle\{ \zeta_\lambda : \lambda^{2^{m+1}-1} 
\geq \epsilon \}\rangle$. 

In the asymmetric case, the multiscale analysis involves a sequence of subspaces 
\begin{equation*}
X_0 \supset X_1 \supset X_2 \supset \cdots  
\end{equation*}
such that $X_0$ is the range of the operator $T$, and for $m \in \ZZ$, $X_m$ represents an 
approximation of the range of $T$ at scale $m$, obtained via the largest singular values and the 
corresponding singular vectors. In particular, let $\{ \sigma_i \}_{i \in \ZZ^+}$ denote the singular 
values of $T$, and $u_{\sigma_i}$ be the corresponding singular vector, we can define 
$X_k:=\langle\{ u_\sigma : \sigma^{2^{k+1}-1} \geq \varepsilon \}\rangle$. Each $W_m$ can then be 
defined as the orthogonal complement of $V_m$ in $V_{m-1}$, as in the original wavelet setting. 
These subspaces $X_k$ can, as in the diffusion wavelet construction, be viewed as the ``low-pass"
portions of the range of $T$, corresponding to the ``approximation" or ``smoothing" subspaces in 
classical wavelet theory, and $W_m$ correspond to the "detail" components. 

One possible disadvantage of this SVD method is that the orthogonalization here is not as careful as the
orthogonalization step in diffusion wavelets \cite{diffwavelets}, and as a result our bases are not
guaranteed to be localized. While this has no apparent adverse effects on the applications to  
examples we experimented, this method may not be suitable for the broader situations where 
DWA applied, particularly where the localization of the scaling functions is crucial. However, in practice 
for Network Embedding datasets, the SVD basis vectors are approximately localized, that is, for any 
given basis vector, the majority of the nodes will be clustered near zero. This ``approximate localization" 
property can be seen in Figure \ref{figure:butterfly}, which shows embeddings based on 2 and 3 singular
vectors. Because this is smaller than the intrinsic dimension of the dataset, most of the nodes take 
values near 0 on the first 3 singular vectors.  For such case, setting a tolerance such that these low 
magnitude values are replaced with zero may improve performance slightly. 

\section{Examples and performance comparisons}\label{sec:numerical}
The following examples illustrate the effectiveness of the method for multi-label classification tasks. In 
this setting, each node is assigned a class label. The goal is to correctly predict the nodes' class labels
based on their embeddings.  In the following experiments, we record for each trial the computational
time, and compare the F1 macro scores to judge the relative quality of each set of label predictions. 
We compare our method to the following existing methods: 
\begin{itemize}
\itemindent=-13pt
\item GraRep \cite{grarep}, which creates multiscale embeddings by  using the SVD of the $k$-step
transition probability matrix to obtain $k$-step representations, and then concatenating the
representations from each scale,
\item DeepWalk \cite{deepwalk}, which uses SkipGram to generate random walk-based embeddings
from unbiased random walks. Here we use the code from the original paper, which uses hierarchical
softmax to speed up computation of the conditional probabilities,
\item Node2vec \cite{node2vec}, a method similar to DeepWalk that uses biased random walks to 
compute embeddings, based on parameters for breadth-first and depth-first sampling,
\item LINE \cite{line}, which uses an edge-sampling algorithm to optimize an objective function that is 
designed to preserve both first and second-order relationships.
\end{itemize}

The Commute Time with SVD embeddings were computed using the algorithm described in Section 3.
The choice of embedding size $d$ for the commute time-based embeddings varies for each dataset,
depending on both the size of the original network and the number of iterations of the compression
algorithms used. For GraRep, LINE, and DeepWalk, all parameters except embedding size $d$ were left
at their default values, and the sizes of the embeddings were chosen to match those of the Commute
Time with Diffusion Wavelet embeddings. The F1 Macro scores reported here are averages based on 10
trials. All computations were performed with an Intel (R) HD Graphics 520 GPU.

The F1 macro scores were computed by generating embeddings based on the data, then using the
embeddings to train a $5$-nearest neighbor classification routine. This classification model was then 
used to predict a class label for each node based on its embedding. F1 scores for each class were 
computed using the formula:
\begin{equation*}
F1_{\text{class } i} = \frac{TP}{TP+\frac{1}{2}(FP + FN)},
\end{equation*}
where $TP$ denotes the number of true positives (nodes which were correctly assigned to class $i$), 
$FP$ denotes the number of false positives (nodes which were correctly assigned to a different class),
and $FN$ denotes the number of false negatives (nodes which should have been assigned to class $i$,
but were not). The overall F1 macro score is the sum over $i$ of all $F1_{\text{class } i}$.

For every dataset, the embeddings created with the method based on the sparse approximation of 
commute times presented in this paper obtained the highest or second-highest F1 macro score. The 
runtime for the method was below or on par with that of methods with comparable F1 scores, with the 
exception of the butterfly dataset. The small size of this dataset allowed the deterministic GraRep 
method to run efficiently and produce accurate embeddings. However, on the larger datasets, our 
method produced higher quality label predictions regardless of network structure (as measured by F1 
scores) compared to GraRep.

\subsection{Butterfly dataset}
We first consider a network from \cite{butterflydata1, butterflydata2} consisting of 832 nodes which
each represent different butterflies.  Each of the organisms belongs to one of 10 species, and between 
55 and 100 pictures of each species were collected from Google Images \cite{butterflydata1}. Edges
indicate similarity between organisms; that is, two nodes will be connected by an edge if their respective
images share features (such as wing color and pattern) in common. The data is available from 
\cite{snapnets}. The commute time embeddings approximated with SVD were based on 5 iterations of
the algorithm, where 50\% of the singular values were retained at each iteration.
The embeddings that led to the best F1 macro scores for the butterfly network were the sparse 
approximation-based commute time embeddings and GraRep (see Table 2). The computation time and
F1 score of the Commute Time with SVD method were comparable to those of the GraRep 
implementation. 

\begin{figure}
\includegraphics[width=0.5\linewidth]{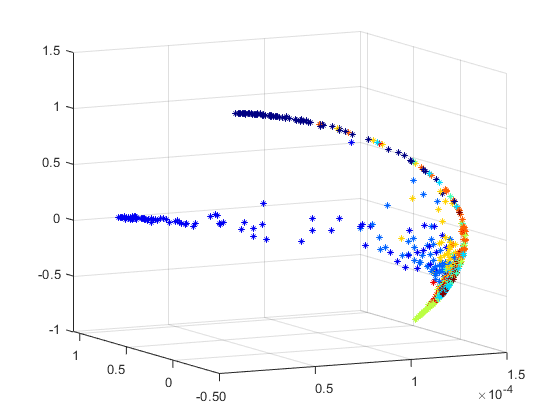}
\hfill
\includegraphics[width=0.5\linewidth]{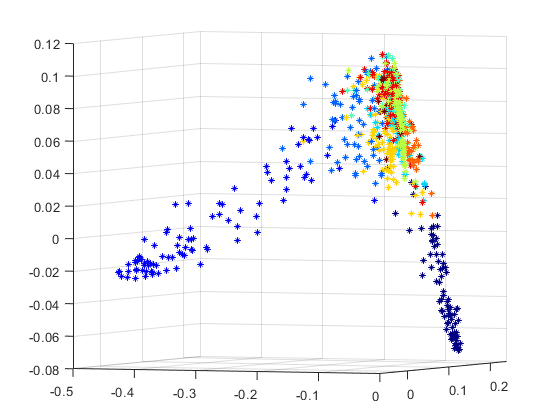}
\caption{ Three-dimensional node embeddings for the butterfly species network, generated using the 
algorithm described in Section 3. Embeddings on the left use sparse approximations produced via
diffusion wavelets, and those on the right use sparse approximations produced via truncated SVD. Each
color represents the species to which that node belongs.}
\label{figure:butterfly}
\end{figure}

Figure \ref{figure:butterfly} plots the three-dimensional embeddings of this network. Each color 
represents the correct classification (i.e., the corresponding butterfly species) of that node. Although 
omitting the rest of the dimensions has decreased the quality of the embeddings (the F1 macro score for 
the three-dimensional embeddings pictured here are only 0.6287 and 0.7490), the clusters 
corresponding to three of the species are visible. Also note that, for these low-dimensional embeddings, 
the localized nature of the embedding vectors has caused the majority of the nodes to be clustered 
around the origin. Slightly higher-dimension embeddings (e.g. $d=27$, as in Table \ref{tab:results}) result 
in more accurate embeddings.

\subsection{Cora dataset}
The Cora dataset \cite{coradata} (available from the Deep Graph Library, \cite{dgl})  is a citation network
consisting of 2708 papers, with edges representing citation relationships. For a given edge, the paper
represented by the source node contained a citation of the target node's paper. The papers are sorted
into seven different topics, corresponding to seven distinct classes. The commute time with SVD
embeddings are based on 4 iterations of this algorithm, retaining 50\% of the singular values at each
iteration.

As shown in Table \ref{tab:results}, for this network, the highest F1 score was achieved by  LINE, with 
node2vec and the SVD-based commute time methods scoring only slightly lower. However, here the
SkipGram methods and LINE had the longest runtimes, while Commute Times with SVD required less
than one minute and still maintained F1 scores near $0.88$. Overall, the sparse approximation-based
commute time  method achieved the highest accuracy at the classification task while computing the
embeddings efficiently.

\subsection{Amazon co-purchase photo dataset}
The Photo segment of the Amazon Co-Purchase network \cite{amazondata},  available from \cite{dgl}, 
consists of 7650 products for sale, each of which is assigned a label representing one of eight distinct
product categories. Edge relationships are determined by which products were recommended when
viewing the product page for a given node. The commute time with SVD embeddings are based on 4 
iterations of this algorithm, where 50\% of the singular values are retained at each iteration.  The 
average F1 macro score over 10 trials was 0.9384.

The Amazon Co-Purchase network is the largest of the networks considered  here, leading to longer 
computational times for all methods. In these trials, the commute time with SVD method and node2vec 
had the best performance in terms of accurate labeling, with F1 scores of 0.9287 and 0.9335 
respectively, but both methods required around 10 minutes. The GraRep implementation was not able to 
accurately classify the nodes. This is likely due to the structure of the network; specifically, the fact that 
the degree matrix of this network, which is inverted in the GraRep algorithm, is singular to machine 
precision. Results for each method are outlined in Table \ref{tab:results}.

\subsection{Email database dataset}
In this example, we consider a network of email communications from a particular organization 
\cite{emaildata1, emaildata2}, available from \cite{snapnets}. This graph contains 1005 nodes, each
corresponding to a member of a certain European research institution. Edges indicate that a particular
pair of members had exchanged emails during the 18-month data collection time period. Each person in 
the network belongs to one of 41 departments. The commute time with SVD embeddings were based on 5 
iterations of the truncated SVD algorithm, where 75\% of the singular values are retained at each 
iteration.
 
For all the methods used, the F1 scores are noticeably lower compared to the other examples, as 
indicated in Table \ref{tab:results}. This may be a result of the structure of the network, i.e., the degree 
matrix for this graph is nearly singular. In particular, this causes GraRep to produce embeddings that
lead to inaccurate classifications, similarly to in the Amazon Co-Purchase example. However, in this
case we still see that the sparse approximation-based commute time embeddings predict the
department labels more accurately than the other methods, while also requiring less time than any other
successful method. 
\begin{table}
\begin{center}
 \caption{Method comparison results from all examples, including runtimes (RT) and F1 macro scores (F1). 
 ``Sparse CT" denotes the commute time embedding using sparse approximations described in Section 3. 
 The embedding dimension $d$ gives the size of the embedding vectors for each example.} 
    \label{tab:results}
\begin{tabular}{lcccr}\toprule
 &  {\textbf{Butterfly}} &  {\textbf{Cora}} &  {\textbf{Amazon}} &  {\textbf{Emails}} \\
Dimension $d$ &   27              &  170        &        480             &     180           \\  
 & RT/F1 & RT/F1  & RT/F1 & RT/F1 \\ 
 \hline
Sparse CT &  0.22 s / 0.9223 & 3.97 s /0.8882 & 105 s / 0.9384 & 0.55 s / 0.6492 \\
DeepWalk & 20 s / 0.8692 & 81 s / 0.8061 & 570 s / 0.9167 & 633 s / 0.5169 \\
node2vec & 183 s / 0.8739 & 240 s / 0.8749 & 570 s / 0.9335 & 240 s / 0.6266 \\
LINE & 210 s / 0.7662 & 150 s / 0.8864 & 178 s / 0.8203 & 141 s / 0.5423 \\
GraRep & 7 s / 0.9103 & 65 s / 0.8333 &  920 s / 0.0451 & 2.145 s / 0.0048 \\
\end{tabular}
\end{center}
\end{table}
\subsection{COVID-19 lung image classifier}
Finally, we apply the methods to a dataset of chest x-rays, saved as 299 $\times$ 299-pixel 
images. The dataset contains x-rays from patients with COVID-19, viral pneumonia patients, and healthy
people, and is available from \cite{lungdataset1, lungdataset2}. In the following experiments we use 60
x-rays from each category. To apply the methods, we reshaped each image into a $299^2$-dimensional
vector, then treated these vectors as the initial representations of each x-ray. We construct the matrix 
$T$ using the Euclidean norm between each two vectors to represent the similarity between those 
nodes. 

For this dataset, we compared the F1 scores, averaged over 10 trials, obtained for this dataset using 
different amounts of training data. Using 90\% of the dataset as training data, the method returns an F1 
macro score of 0.7954 on average, while reducing the dimension of the vector representations from 
$299^2$ to 3. As the amount of training data is decreased, the F1 macro score also decreases, but even
when training data is decreased to $10\%$, the average F1 score remains above 0.5, at 0.5372. 
The commute time with SVD embeddings were based on 6 iterations of this algorithm, where 50\% of 
the singular values are retained at each iteration. 
 
\begin{figure}
\includegraphics[width=0.5\linewidth]{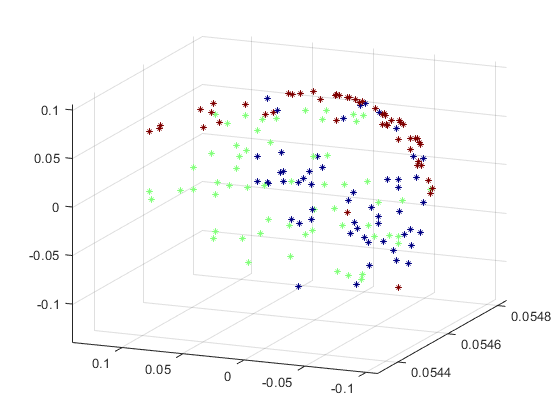}%
\hfill
\includegraphics[width=0.5\linewidth]{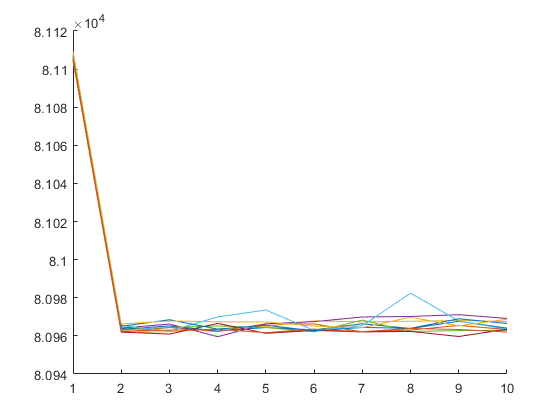}
\caption{(a): Three-dimensional node embeddings for the chest x-ray classifier, generated via the proposed method. 
Each color represents the diagnosis (normal, COVID-19, or viral pneumonia) associated with that x-ray.
(b):  For each of 10 trials, the cross-entropy loss over 10 iterations of SGD is recorded.}
\label{figure:xray}
\end{figure}

Figure \ref{figure:xray} shows the method's efficacy for clustering. The low-dimensional embeddings 
show distinct grouping by color such that red points represent images taken from patients with 
COVID-19, light blue represents images of healthy patients, and dark blue represents images from
patients with viral pneumonia. Figure \ref{figure:xray} (b) shows the decrease in cross entropy loss over 
10 iterations of SGD, for 10 different trials. In all trials, cross-entropy loss drops steeply for the first steps
of gradient descent, quickly converging to low values within 2 iterations.   

\section{Analysis of the scheme}
In the following section, we will address some theoretical issues justifying the proposed method to shed 
light on further improvements. 
Specifically, we'll elaborate on the performance comparison between DWA and SVD for sparse approximation, 
present an applicable form of the Johnson-Lindenstrauss lemma, and discuss the relationship between the 
scheme presented here and the diffusion maps \cite{coifman}. 

\subsection{More on DWA vs SVD}
We first want to show that the SVD can achieve at least the same accuracy as DWA for sparse 
approximations of transition matrices on networks. Our discussion is based on the following definition of 
$\varepsilon$-span from \cite{diffwavelets}:
\begin{defn}
Let $\mathcal{H}$ be a Hilbert space such that $\{w_j\}_{j \in J} \subset \mathcal{H}$.
A set of vectors $\{v_i\}_{i \in I}$ $\varepsilon$\textbf{-spans} the set of vectors $\{w_j\}_{j \in J}$ if 
\begin{equation*}
||P_{\langle \{v_i\}_{i \in I} \rangle}w_j-w_j||_{\mathcal{H}} \leq \varepsilon,
\end{equation*}
where $P_{\langle \{v_i\}_{i \in I} \rangle}$ represents orthogonal projection onto the span of $\{v_i\}_{i \in I}$. 
We also say that $\langle \{v_i\}_{i \in I} \rangle$ is an $\varepsilon$-span of $\langle \{w_j\}_{j \in J} \rangle$.
\end{defn}
If the sparse approximations $T_k$ are computed using DWA, then the $T_k \approx_\varepsilon T^{2^k}$  in 
the sense that $T_k$ is expressed in a basis that $\varepsilon$-spans the columns of $T_{k-1}$. 
Alternatively, the truncated SVD-based approximation, $T_k=U_{k-1} T_{k-1}^2 U_{k-1}^T$ where $j$ singular 
values were retained, is an approximation of $T^{2^k}$ in the sense that the truncated SVD is an approximation 
of $T_{k-1}$ up to the next largest singular value, $\sigma_{j+1}$:
\begin{equation*}
|| T_{k-1} - U_{k-1} \Sigma_{k-1} V_{k-1}^T ||_2 = \sigma_{j+1}.
\end{equation*}
It is well known that this is the best possible $j$-rank approximation of $T_{k-1}$. In the self-adjoint case, using 
this ideal $j$-rank approximation leads to 
\begin{equation*}
T_{k-1}^2 = T_{k-1}^T T_{k-1} = U_{k-1} \Sigma_{k-1} V^T_{k-1} V_{k-1} \Sigma_{k-1} U_{k-1}^T = U_{k-1} 
\Sigma_{k-1}^2 U_{k-1}^T.
\end{equation*}
In the non-self-adjoint case, we can express $T_{k-1}^2$ in terms of $U_{k-1}$ similarly via the change of basis  
\eqref{eq:chain}. Further, we would expect that this SVD approximation results in less error, compared to a diffusion 
wavelet compressed representation of the same dimension. Theorem \ref{thm:svd} formalizes this result. 
Proof of the following theorem can be found in the Appendix. 

\begin{lemma}
$U_k$ is a $\sigma_{j+1}$-span of the columns of $T_{k}$,  that is, for any vector 
in the span of the columns of $T_{k}$, its orthogonal projection onto $span(U_k)$ is within $\sigma_{j+1}$ of
the original vector. 
\end{lemma}
Recall that we have assumed the transition matrix  $T$  is a local operator with its columns having small support.
\begin{theorem}
Suppose $T$ is either symmetric or invertible. Let $\Phi_\ell$ denote the diffusion wavelet basis at the $\ell$th 
compression level, constructed with precision $\varepsilon$,  with number of columns being $j_1$.
Then, it is possible to choose $j_2 \leq j_1$ such that  the first $j_2$ left singular vectors of $T_k$, denoted 
$U_k$, $\varepsilon_2$-span the range of $T_k$ with $\varepsilon_2 \leq \varepsilon$.  
\label{thm:svd} 
\end{theorem}

Furthermore, the SVD basis performs with greater efficiency on a broader class of operators $T$. At worst, both 
the full SVD algorithm and the diffusion wavelet algorithm can have complexity up to $O(n^3)$ on a general 
$n \times n$ matrix. In order for diffusion wavelets to perform with computational complexity better than 
$O(n^3)$, $T$ must have $\gamma$-strong decay (for some $\gamma > 0$, if for all $\lambda \in (0,1)$, 
eigenvalues $\lambda_k$ satisfy $\#\{  k: \lambda_k \geq \lambda  \}\leq C \log_2^\gamma/{\lambda}$)
\cite{diffwavelets} with $\gamma$ sufficiently large and $\varepsilon$ not too small. 

However, if we assume that this $\gamma$-decay holds for $T$, its singular values will also decay fast. The SVD 
algorithm, then, will require fewer singular vectors to obtain the same accuracy. The truncated SVD of a low-rank 
$m \times n$ matrix with rank $k$ can be computed with complexity $O( mnk)$, and when sufficiently few singular 
values are needed, fast algorithms exist for sparse matrices (e.g. Lanczos bidiagonalization \cite{lanczos}) where 
complexity depends linearly on the number of nonzero elements in $T$, rather than $n$. This difference in 
performance can be seen empirically in Table \ref{tab:svdcomparison}, which compares the two methods on the 
butterfly dataset. A similar difference in computational time was also observed for the other datasets presented 
in the following section.

\subsection{ Modified Johnson-Lindenstrauss Lemma}
To analyze Green function-based commute time embeddings from a more fundamental perspective of the 
Johnson-Lindenstrauss Lemma. In particular, the Euclidean distances between these embeddings 
approximate up to $\varepsilon$ the commute time distance (or, for diffusion wavelet-based embeddings, 
the square root of the commute time distance) between the original nodes.

By a straightforward application of the Johnson-Lindenstrauss Lemma, the commute time can be 
approximated with high probability by the product of $G^{1/2}$ and a constant multiple of a random matrix with 
normally distributed entries. Let $\Phi$ be the basis of scaling functions obtained by  applying DWA to the matrix 
$vol\cdot G D^{-1}$ with precision constant $\varepsilon$, and $\Phi(x):=\Phi^T\delta_x$ denotes the $x$th row of 
$\Phi$, and the embedding for node $x$, then for all $x,y \in S$,
\begin{align}
- \varepsilon + \sqrt{CT(x,y)} \leq || \Phi^T(\delta_x-\delta_y)|| \leq \varepsilon + \sqrt{CT(x,y)} .
\end{align} 

Similarly, let $\tilde{U} \Sigma \tilde{U}^T \approx vol\cdot G D^{-1}$ represent the truncated Cholesky 
decomposition of $vol\cdot G D^{-1}$ using the $\alpha N$ leading singular values 
$\sigma_1\geq...\geq\sigma_{\alpha N}$, for some $\alpha \leq 1$. Let $\Theta$ denote the embedding matrix 
obtained from the leading left singular vectors $\tilde{U}$, such that $\Theta = \Sigma^{1/2} \tilde{U}$ and 
$\varepsilon = O(\sigma_{\alpha N+1})$, then for all $x,y \in S$,
\begin{align}
- \varepsilon + CT(x,y) \leq || \Theta^T(\delta_x-\delta_y)|| \leq \varepsilon + CT(x,y) .
\end{align}
These results are simple to obtain from the definition of $\varepsilon$-span and the fact that 
$CT(x,y)=|| (vol \cdot G D^{-1})^{1/2}\delta_x - (vol \cdot G D^{-1})^{1/2}\delta_y ||_2$.

\subsection{ Relationship with diffusion maps}
Finally, we show that embeddings given by diffusion wavelet scaling functions are equivalent up to a rotation to the 
embeddings produced from any given kernel eigenmap method. In particular, this means the embeddings produced
via the application of diffusion wavelets to the matrix $vol \cdot GD^{-1}$ are equivalent up to a rotation to the 
embeddings given by kernel PCA on $vol \cdot GD^{-1}$, and so they are also equivalent up to a rotation to the 
commute time embeddings from \cite{qiu}.

First, we begin by summarizing the connection between the diffusion map and other kernel eigenmap methods. 
In \cite{coifman}, it is shown that any kernel eigenmap method can be viewed as solving 
\[
\min_{Q_2(f)=1} Q_1(f)\text{, where } Q_1(f)=\sum_x Q_x(f),
\]
where $Q_2, Q_x$ are symmetric positive semi-definite quadratic forms with $Q_x$ local. 
For example, for a kernel $k$, we can take 
\begin{align*}
Q_1(f) &= \sum_x \sum_y k(x,y) (f(x)-f(y))^2 \\
Q_2(f) &= \sum_x v(x)f(x)^2.
\end{align*}
Then, $Q_2^{-1}Q_1$ represents a discretization of a differential operator (i.e., a Laplacian). 

\begin{theorem}
Suppose we have a kernel-based method (e.g. $Q_1, Q_2$ satisfying the properties described above).
Then the embeddings obtained from the diffusion wavelet scaling functions for $Q_2^{-1}Q_1$ at level $j$ are 
equivalent, up to a rotation, to the $p_j$-dimensional embeddings obtained from the diffusion maps embedding, 
where $p_j$ gives the number of scaling functions at scale $j$.
\end{theorem}

The above theorem and its proof (provided in the Appendix) are a slight generalization of  results from
\cite{mahadevan}. In particular, since the commute time embeddings in \cite{qiu} are equivalent to kernel PCA 
on the matrix $vol \cdot GD^{-1}$, we can apply the above theorem with $Q_2^{-1}Q_1 = vol \cdot GD^{-1}$ to 
see that the embeddings  produced via the diffusion wavelet method are equivalent up to a rotation to the 
$p_j$-dimensional commute time embeddings, and thus these embeddings will also preserve commute times 
between nodes.

\section{Conclusion} 
This paper outlined a new method for embedding the vertices of a network while preserving commute time 
distances between nodes and optimizing cross-entropy loss. Future work could include a more complete analysis 
of the method, and possibly further improvements to the algorithm. In particular, it is likely to improve upon the 
the SGD stage that makes the largest contribution to computation time which may involve incorporating 
randomized algorithms to more efficiently optimize cross-entropy loss.

While this article focuses on undirected networks, this method is applicable for the directed case as well.
For a directed graph, the adjacency matrix $A$, as well as the random walk Laplacian $L$ will be 
asymmetric, which means $L$ may not be diagonalizable, and the definition of the Green function 
need to be make use of the Moore-Pembrose pseudoinverse, $L^+ = (L^TL)^{-1}L^T$. Since $L^+$ is still 
the left inverse of $L$, and the method presented here does not require explicit computation of eigenvalues or 
eigenvectors of $L$, the computation of the embeddings can proceed. More details will be provided in future 
expositions.

\appendix\section{Proofs of Theorems}
\subsection{Lemma 5.1}
\begin{proof}
Let $v \in span(T)$, and let $T=U \Sigma V^T$ denote the SVD. Then, there exist some vector $ w, ||w||_2 =1$, 
such that $v = Tw = U\Sigma V^T w$. Let the orthogonal projection of $v$ onto $U_1$ be denoted $\tilde{v}$. 
To prove that $U_1 = \begin{bmatrix}
u_1 & u_2 & \cdots & u_j
\end{bmatrix}$ is an $\sigma_{j+1}$-span of the columns of $T$, we need to show that
\[
||v-\tilde{v}||_2 \leq \sigma_{j+1}.
\]
Since the columns of $U_1$ are orthonormal, we have:
\begin{align*}
||v-\tilde{v}||_2 &= || U \Sigma V^Tw- U_1 U_1^T U \Sigma V^T w||_2 \\
&= || U \Sigma V^Tw- U_1 \Sigma V^T w||_2 \\
&\leq \sigma_{j+1},
\end{align*}
where the second line holds because 
\[
U_1^T U = \begin{bmatrix}
I_{j \times j} & 0 \\
0 & 0 \\
\end{bmatrix}.
\]
Hence $U_1$ is a $\sigma_{j+1}$-span of $T$, by definition.
\end{proof}

\subsection{Theorem 5.2}
\begin{proof}
Fix $\varepsilon >0$. 
Choose $j_1$ such that $\sigma_{j_1} > \varepsilon$ and $\sigma_{j_1+1}\leq \varepsilon$. (Note that such $j_1$ can be assumed to exist without loss of generality, because if $\varepsilon$ is so small that no such $j_1$ exists, then applying DWA with this $\varepsilon$ will not reduce the dimension, and $\Phi_1$ will have the same number of columns as $T$. In this case, $j_1=N$.)

\textbf{Symmetric Case:}
In the case where $T$ is symmetric, since the eigenvalues of $T$ in this setting are nonnegative, $\sigma_{j+1} = \lambda_{j+1}$ for all $j$. 
Applying DWA to T, then, results in a basis $\Phi_1$ spanning the space
\[V_1 = span(\{ \zeta_\lambda : \lambda \geq \varepsilon   \}),\] 
where the $\zeta_\lambda$ denote eigenvectors.
Choosing $j_1 = \inf \{j: \sigma_{j+1}\leq \varepsilon \}$, and defining $U_1$ to be the first $j_1$ left singular vectors of $T$, we have $U_1 = \{ u_m : \sigma_m \geq \varepsilon \} = \{ \zeta_\lambda : \lambda \geq \varepsilon\}$. 

After the first compression step, we have from diffusion wavelets \\ $T_1^{DWA} = \Phi_1 T^2 \Phi_1^T$, 
and using SVD, $T_1^{SVD} = U_1 T^2 U_1^T$, where $\Phi_1$ and $U_1$ are two bases $\varepsilon$-spanning $V_1$. 
We also have:
\begin{align*}
T_1^{SVD} &= U_1 T^2 U_1^T 
= U_1 U\Sigma U^T U \Sigma U^T U_1^T 
= U_1 U \Sigma^2 U^T U_1^T, 
\end{align*}
which implies that, if we denote the singular values of $T_1^{SVD}$ as $\sigma^{T_1}$,
\[
\{ m: \sigma^{t_1}_{m+1} \leq \varepsilon \} = \{ m: (\sigma^{2}_{m+1})^{t_1/2} \leq \varepsilon \}
= \{ m: (\sigma^{T_1^{SVD}}_{m+1})^{t_1/2} \leq \varepsilon \}.
\]

Let $t_k := 2^{k+1}-1$. Then, defining $V_2 = span(\{ \zeta_\lambda : \lambda^{t_1} \geq \varepsilon   \})$,  $j_2 = \inf \{ m: (\sigma^{T_1^{SVD}}_{m+1})^{t_1/2} \leq \varepsilon \}$, and $U_2$ as the first $j_2$ singular vectors of $T_1^{SVD}$, clearly $dim(V_2)=j_2$. So both $\Phi_2$ and $U_2$ will again have the same dimension, and both bases will $2\varepsilon$-span the range of $T^2$.  

The same argument holds for the general $k$th step, where $V_k = span(\{ \zeta_\lambda : \lambda^{t_{k-1}} \geq \varepsilon   \})$ and $j_k = \inf \{ m: (\sigma^{T_{k-1}^{SVD}}_{m+1})^{t_{k-1}/2} \leq \varepsilon \}$.
Since the eigenvector basis for $V_k$ and the leading singular vectors $U_k$ are identical in the symmetric case, clearly the bases $\Phi_k$ and $U_k$ span the same space, and therefore have the same dimension. By construction, both $T^{SVD}_k$ and $T^{DWA}_k$ are $\varepsilon$-approximations of $T^{2^k}$. So the given choices of $j_k$ identify a basis of singular vectors of equal or lesser dimension to $\Phi_k$, that also $\varepsilon$-span the same space, and the theorem is satisfied. 

\textbf{Asymmetric case:}
 If $T$ is asymmetric, we can symmetrize for diffusion wavelets by multiplying by $T$ by its adjoint. (Note that, if $T = U \Sigma V^T$, $TT^T = U\Sigma^2U^T$, so the eigenvalues of $TT^T$ are the squares of the singular values of $T$, and the eigenvectors of $TT^T$ are the left singular vectors of $T$.) If we instead apply diffusion wavelets to $T$ itself, $V_k$ can instead be defined as the span of the basis vectors themselves, or as an $\varepsilon$-span of the columnspace of $T$. Since $T$ and $TT^T$ have identical ranges so long as $T$ is invertible, the diffusion wavelet basis obtained from $TT^T$ must also be a $\varepsilon$-span of $V_1= span(\{ \zeta_\lambda : \lambda \geq \varepsilon   \})$, so it suffices to compare the SVD subspace to the subspace defined by $TT^T$. 
 
For nonsingular $T$,  the diffusion wavelet scaling functions $\varepsilon$-spanning the range of $T$ must also $\varepsilon$-span the range of $TT^T$, and therefore 
\[
span(\Phi_1) \approx_\varepsilon V_{1,TT^T}.
\] 

Let $u_j, \sigma_j, j \in \ZZ$ denote singular vectors and values of $T$, and $\zeta_{\lambda_j}, \lambda_j, j \in \ZZ$ denote eigenvectors and eigenvalues of $TT^T$. Then
\begin{align*}
V_{1, TT^T} &= span(\{ \zeta_\lambda : \lambda \geq \varepsilon   \}) = span(\{ \zeta_\lambda : \lambda = \sigma^2, \sigma^2 \geq \varepsilon   \} ) = span(\{ u_j: \sigma_j^2 \geq \varepsilon  \}), \\
\end{align*}
Note that $\{ u_j: \sigma_j^2 \geq \varepsilon  \} \subset \{ u_j: \sigma_j \geq \varepsilon  \}$. By choosing $j_k = \inf \{m: \sigma_{m+1}^2 \leq \varepsilon \}$, the basis $U_1$ consisting of the first $j_k$ left singular vectors will contain at most as many vectors as the diffusion wavelet basis $\Phi_1$ on $T$.
We now have a subset of the $j_k$ leading singular vectors of $T$ that spans a low-pass portion of the range of $T$.  

Likewise, for the $k$th compression level,
\begin{align*}
V_{k,TT^T} &= span(\{ \zeta_\lambda : \lambda^{t_k} \geq \varepsilon   \}) \\
&=span( \{ u_\sigma : (\sigma^2)^{t_k} \geq \varepsilon \}) \\
&= span ( \{ u_\sigma : \sigma^{t_{k+1}-1} \geq \varepsilon \}), 
\end{align*}
and our basis of singular vectors $U_k$ will $\varepsilon$-span the same space as $\Phi_k$ when we choose $j_k = \inf \{ m: \sigma^{t_{k+1}-1} \geq \varepsilon\}$. As in the symmetric case, we can instead define
$j_k = \inf \{ m: (\sigma^{T_{k-1}}_{m+1})^{(t_{k+1}-1)/2^{k-1}}\}$, and use the SVD of $T_{k-1}$ to compute $T_k$.

In either the asymmetric or symmetric cases, there is some subset of left singular vectors of $T_k$ that spans an equivalent (up to $\varepsilon$) space to the diffusion wavelets basis at a particular level. By the previous lemma, since the basis $U_k$ is a $\sigma_{j+1}$-span of the range of $T_k$ with $\sigma_{j+1} \leq \varepsilon$, the approximation using this basis is guaranteed to provide an equal or better error bound compared to the diffusion wavelet basis. Additionally, by the choice of $j_k$ and the compressed nature of the $T_k$, we can expect the SVD bases to contain a smaller, or at worst equal, number of vectors. Taking $j_k$  as defined above for the $k$th compression step, then, leads to a basis of singular vectors that satisfies the theorem. 
\end{proof} 

\subsection{Theorem 6.3}
\begin{proof}
The diffusion maps algorithm embeds nodes by solving the eigenproblem 
\[
Q_2^{-1}Q_1f=\lambda f,
\]
and using $\Psi_t(x) := (\lambda_1^t \psi_1(x) \hspace{.5cm} \lambda_2^t \psi_2(x) \cdots \lambda_{s(\delta,t)}^t \psi_{s(\delta,t)}(x) )^T$ where $s = \max\{ l \in \mathbb{N} \text{ such that } |\lambda_l|^t>\delta|\lambda_1|^t \}$ and $\lambda_k$, $\psi_k$ are the eigenvalues and eigenvectors of the transition matrix $P$. We order the $\lambda_k$ such that $1=\lambda_0 > |\lambda_1 |\geq \cdots$. 
The operator $Q_2^{-1}Q_1$ is a discretization of some differential operator, and in particular, it is an infinitesimal generator of the random walk if its eigenvalues are between 0 and 1 (see \cite{coifman} for further discussion of this operator). Then, $P$ and $Q_2^{-1}Q_1$ have the same eigenvectors, and the leading eigenvectors of one correspond to the last eigenvectors of the other.

Let $j = \min\{k: \lambda_0^{t_k} > \varepsilon \}$. Define $V_{1:p_j}$ as the matrix whose columns are the first $p_j$ eigenvectors of $Q_2^{-1}Q_1$, and let  $[ \phi_j ]_{\phi_0}$ be the matrix of scaling functions of $Q_2^{-1}Q_1$ at level $j$. The scaling functions of $Q_2^{-1}Q_1$ at level $j$, by construction, give (up to $\varepsilon$) a localized basis for $V_j = \langle \psi_{\lambda} : \lambda \in \sigma(T), \lambda^{t_j} \geq \varepsilon \rangle$. This $V_j$ is  the space spanned by the $p_j$ eigenvectors associated with the largest eigenvalues of $P$, or equivalently, the eigenvectors associated to the $p_j$ least eigenvalues of $Q_2^{-1}Q_1$.

Note that the columns of $V_{1:p_j}$ and $[ \phi_j ]_{\phi_0}$ span the same space. 
Hence
\begin{align*}
V_{1:p_j}^TV_{1:p_j} &= I = [ \phi_j ]_{\phi_0}^T[ \phi_j ]_{\phi_0}, \\
\Rightarrow V_{1:p_j} &= V_{1:p_j}V_{1:p_j}^TV_{1:p_j} = [ \phi_j ]_{\phi_0}([ \phi_j ]_{\phi_0}^T V_{1:p_j}) 
\end{align*}
Call this last matrix in parentheses $Q$. Then $QQ^T = [ \phi_j ]_{\phi_0}^T V_{1:p_j}V_{1:p_j}^T[ \phi_j ]_{\phi_0} = I$, and likewise $Q^TQ = I$, so $Q$ is a rotation matrix. Therefore, the $p_j$-dimensional diffusion map embeddings are equivalent up to a rotation to the $j$th-level scaling function-based embeddings.
\end{proof}

\bibliographystyle{siamplain}
\bibliography{commutetimessvd}
\end{document}

%% file: sparseapprox_shared.tex
\usepackage{lipsum}
\usepackage{amsfonts}
\usepackage{graphicx}
\usepackage{epstopdf}
\usepackage{algorithmic}
\ifpdf
  \DeclareGraphicsExtensions{.eps,.pdf,.png,.jpg}
\else
  \DeclareGraphicsExtensions{.eps}
\fi

\newsiamremark{remark}{Remark}
\newsiamremark{hypothesis}{Hypothesis}
\crefname{hypothesis}{Hypothesis}{Hypotheses}
\newsiamthm{claim}{Claim}

\headers{Network Embedding Using Sparse Approximation of Random Walks}{P. Mercurio and D. Liu}

\title{Network Embedding Using Sparse Approximation of Random Walks
\thanks{Submitted to the editors August 25th, 2023.
\funding{This work was funded by NSF-DMS 1720002.}}}

\author{Paula Mercurio\thanks{Department of Mathematics, St. Olaf College, Northfield, MN
  (\email{mercur1@stolaf.edu}).}
\and Di Liu\thanks{Department of Mathematics, Michigan State University, East Lansing, MI
  (\email{liudi1@msu.edu}).}
}

\usepackage{amsopn}
\DeclareMathOperator{\diag}{diag}
